\documentclass[letterpaper, 10 pt, conference]{ieeeconf}  

\IEEEoverridecommandlockouts                              

\usepackage{cite}
\usepackage{amsmath,amssymb,amsfonts}
\usepackage{algorithmic}
\usepackage{graphicx}
\usepackage{textcomp}
\usepackage{xcolor}
\usepackage{hyperref}
\usepackage{subcaption}

\title{\LARGE \bf
E-VFIA : Event-Based Video Frame Interpolation with Attention
}

\author{Onur Selim Kılıç, Ahmet Akman and A. Aydın Alatan
\thanks{All authors are members of Electrical and Electronics Engineering Department and Center for Image Analysis (OGAM) in 
        Middle East Technical University (METU), 06800 Ankara, Turkey
        {\tt\small \{selimk,ahmet.akman,alatan\}@metu.edu.tr}}%
}

\begin{document}

\maketitle
\thispagestyle{empty}
\pagestyle{empty}

\begin{abstract}

Video frame interpolation (VFI) is a fundamental vision task that aims to synthesize several frames between two consecutive original video images. Most algorithms aim to accomplish VFI by using only keyframes, which is an ill-posed problem since the keyframes usually do not yield any accurate precision about the trajectories of the objects in the scene. On the other hand, event-based cameras provide more precise information between the keyframes of a video. Some recent state-of-the-art event-based methods approach this problem by utilizing event data for better optical flow estimation to interpolate for video frame by warping. Nonetheless, those methods heavily suffer from the ghosting effect. On the other hand, some of kernel-based VFI methods that only use frames as input, have shown that deformable convolutions, when backed up with transformers, can be a reliable way of dealing with long-range dependencies. We propose event-based video frame interpolation with attention (E-VFIA), as a lightweight kernel-based method. E-VFIA fuses event information with standard video frames by deformable convolutions to generate high quality interpolated frames. The proposed method represents events with high temporal resolution and uses a multi-head self-attention mechanism to better encode event-based information, while being less vulnerable to blurring and ghosting artifacts; thus, generating crispier frames. The simulation results show that the proposed technique outperforms current state-of-the-art methods (both frame and event-based) with a significantly smaller model size.

\end{abstract}

\paragraph*{Multimedia material} The code is available at \url{https://github.com/ahmetakman/E-VFIA} 

\section{INTRODUCTION}In robotics applications, where fast-moving agents are involved, high frame-rate video streams are necessary for agile reaction of control systems. There has been dedicated hardware for capturing high frame-rate videos; however, they are quite expensive. Therefore, increasing the frame rate with the involvement of additional processing yields more affordable high frame-rate videos and smoother slow-motion videos.

Video frame interpolation (VFI) is a vision task where a non-captured image sample is inserted between consecutive frames to increase the rate of low frame-rate videos. Since the movements appearing on the scene can be complex and their displacements are large, video frame interpolation is mostly considered as a challenging task. 

The state-of-the-art methods that only use frames aim to estimate motion from consecutive images accurately. Such frame-only methods are divided into two mainstream categories, as (i) flow-based methods \cite{DAIN, BMBC, qvi_nips19,sim2021xvfi, Lu_2022_CVPR,huang2022rife}, and (ii) kernel-based methods \cite{Shi_2022_CVPR, Lee2020AdaCoFAC, Niklaus_CVPR_2017, Niklaus2021}. 

Although frame-only approaches are regarded as precise when relatively short trajectories are involved, they might fail to estimate complex movements of fast-moving objects. The main reason for these algorithms fail in situations with fast-moving objects is due to the fact that they can only make linear predictions for the trajectories between keyframes to model the complex motion of the objects. Recent advances in mobile camera technologies lead more affordable (relative to the dedicated hardware) devices to have relatively higher frame-rate video recording capability, but this is also limited by the memory requirements of the devices, making it impossible to record long sequences of high frame-rate video by such hardware. Thus, the use of VFI methods still offer memory and storage efficiency.

Event-based cameras are novel vision sensors that detect only pixel-level brightness changes. The pixels in the sensor array asynchronously output information at high speeds (up to 1-$\mu s$ precision \cite{Survey}) for a high dynamic range (up to 120 dB \cite{Survey}). Therefore, event-based cameras are suitable sensors for obtaining additional information related to the moving objects in the scene. On the other hand, the asynchronous and fast nature of the event sensor implicates low data bit-rate due to the limited number of event data, while agile sensing is possible. Therefore, they provide beneficial information in temporal information that applies to VFI problem.

Recent studies \cite{Tulyakov_2022_CVPR} \cite{Tulyakov_2021_CVPR} argue that using additional information from the event-based sensors might outperform frame-only interpolation methods for VFI. The methods in \cite{Tulyakov_2022_CVPR} and \cite{Tulyakov_2021_CVPR} are follow-up studies of each other that estimate intermediate frames by warping the images after estimating the motion displacements by the help of events. The aforementioned algorithms have primarily been built on a combination of specific purpose-trained hourglass networks, which are computationally expensive due to the high number of parameters. Based on our preliminary simulations, the interpolated frames in \cite{Tulyakov_2022_CVPR} and \cite{Tulyakov_2021_CVPR} suffer from ghosting and blurring artifacts.

Transformers \cite{vaswani2017attention} are novel neural structures initially developed to model long-range dependencies in natural language processing (NLP) tasks. Researchers took inspiration from the success of the transformers in NLP and applied a similar approach to various vision tasks \cite{dosovitskiy2020image}. The idea of using transformers has been employed recently for frame-only VFI tasks both in kernel-based \cite{Shi_2022_CVPR} and flow-based methods \cite{Lu_2022_CVPR}. However, they also perform poorly in case of fast-moving objects present in the scene. Aforementioned studies inspire our work for modeling long-term dependencies using transformer structures in a lighter manner.

\textbf{Motivation.} The algorithms proposed in \cite{Lu_2022_CVPR}, \cite{Shi_2022_CVPR}, and \cite{Lee_2020_CVPR} mainly argue that if the movements of the visual content in consecutive images is encoded by a network and fused with the real images, one can achieve accurate generation of the intermediate frame as long as the optical flow modeling is accurate enough. The fact that frame-only methods suffer from modeling of complex and fast motions leads to use of additional sensor information. The event cameras might be an excellent way to capture more accurate temporal information. Additionally, the success of \textit{deformable convolutions} \cite{Deformable} for synthesizing intermediate frames motivates for a kernel-based video frame interpolation method, which should be based upon deformable convolution concept and backed up with an attention mechanism. In this work, we propose an attention-based, light-weight VFI network, namely \textit{event-based video frame interpolation with attention} (E-VFIA), that outperforms the visual quality of the current SoA event-based VFI methods in BS-ERGB dataset \cite{Tulyakov_2022_CVPR}.

The main contributions of this work can be summarized as follows:

\begin{itemize}
    \item We propose E-VFIA, the first kernel-based algorithm to utilize deformable convolutions to fuse event-based information and standard images for video frame interpolation with events. 
    \item E-VFIA achieves significant improvement (up to $1.04dB$) against the state-of-the-art methods that use only key-frames and events together with key-frames. The proposed method achieves this promising result by approximately $2.07$ million parameters which is much less than its counterparts, making it more suitable for mobile robotics applications.
    \item We have utilized voxel grids to represent events. In order to extract the temporal information from the events efficiently, we analyze the effect of voxel grid size on the VFI performance, and it is observed that using voxel grids with higher temporal resolutions improves performance.
    \item We utilize both temporal and spatial pooling operations to associate fast-moving objects between consecutive images. By the help of such pooling operations, the objects that are moving fast are predicted better, since utilization of images at different resolutions enables data aggregation between different ends of the image.
\end{itemize}

\section{Related Work}
\textbf{Video Frame Interpolation.} VFI is a well-studied problem with mature solutions. The recent learning based methods can be divided into three groups: flow-based \cite{DAIN, BMBC, qvi_nips19,sim2021xvfi, Lu_2022_CVPR,huang2022rife}, kernel-based \cite{Shi_2022_CVPR, Lee2020AdaCoFAC, Niklaus_CVPR_2017, Niklaus2021}, and phase-based \cite{meyer2018phasenet}.

\emph{Flow-based approaches.} Essentially, flow-based methods \cite{DAIN, BMBC, qvi_nips19,sim2021xvfi, Lu_2022_CVPR,huang2022rife} estimate intermediate flow from left and right frames. Flow-based methods assume the motion in the scene is linear in its trajectory. Common architectural selections are mostly a variation of hourglass backbones; hence their performance is limited in occlusions, nonlinear trajectories, and brightness changes. 

\emph{Kernel-based approaches.} Kernel-based methods \cite{Shi_2022_CVPR, Lee2020AdaCoFAC, Niklaus_CVPR_2017, Niklaus2021} aim to estimate a series of convolutional kernels that model the movements of the image patches implicitly. However, they might fail to model movements larger than the kernel size. A recent study \cite{Lee2020AdaCoFAC} enables kernel-based methods to model large motions by the development of deformable convolutions concept \cite{Deformable}. A recent promising approach \cite{Shi_2022_CVPR} extends this concept by utilizing an encoder-decoder mechanism and aims to model the long-term dependencies.

\emph{Phase-based approaches.} As indicated in \cite{Lee2020AdaCoFAC}, the method in \cite{meyer2018phasenet} takes video frames as linear combinations of different waveforms with different directions and frequencies. This approach interpolates phase and magnitude of each band of wavelet transform. This method is considered effective and efficient in both performance and runtime. Unfortunately, it is still constrained with large displacements, especially for high-frequency components.

\textbf{Utilization of Additional Sensors.} The idea of using additional information coming from an auxiliary camera has been investigated in the literature by low resolution high frame-rate cameras \cite{Paliwal_2020,gupta:2009}. It should be noted that the temporal resolutions of event cameras are significantly higher with respect to affordable low resolution high frame-rate cameras. 

\textbf{Event-Based Approaches.} Event cameras offer higher dynamic range capability in addition to low data-rate and high temporal information flow. Thus, exploiting the properties of the event camera can be considered as a novel direction to investigate for VFI. Some studies \cite{Tulyakov_2021_CVPR,Tulyakov_2022_CVPR,DeblurringECCV2020 ,WeaklySupervised,EvIntSR,Wang_2021_ICCV} aim to overcome the VFI problem by using the information from event-based cameras. In a promising approach, \cite{Tulyakov_2021_CVPR} flow masks for intermediate frames from event voxels are estimated. Then, the algorithm generates artificial frames by warping the adjacent images by the estimated flow. Finally, an attention averaging module is used for fusing three sister images obtained from warping refinement module. In a two stage method \cite{WeaklySupervised}, the network fuses events with images in the first stage, whereas in the next part, the fused vectors are evaluated and integrated with a subpixel transformer network. The results in \cite{Tulyakov_2021_CVPR} are improved further by introducing a motion spline estimator, and multiscale feature fusion modules \cite{Tulyakov_2022_CVPR}. The motion spline estimator enables the pipeline to take into account the continuous nature of the movements on the scene. Since this network \cite{Tulyakov_2022_CVPR} has approximately 70M parameters for encoding the events to voxels, the number of time bins is limited in terms of memory. Such an approach leads event representation to be more blurry.

This paper addresses the primary limitations mentioned in \cite{Tulyakov_2022_CVPR} and \cite{Kalluri2020FLAVRFV}. We extend the approach in \cite{Shi_2022_CVPR} by using both events and frames. Since the event information already represents valuable details, we adopt a simpler input stage with a smaller attention mechanism, significantly reducing the number of parameters. The proposed method is relatively lightweight and can represent events in higher temporal resolution with deeper voxel grids while achieving improved interpolation results. 

\section{Proposed Method}

\begin{figure*}[htbp]
    \centering{\includegraphics[width=\linewidth]{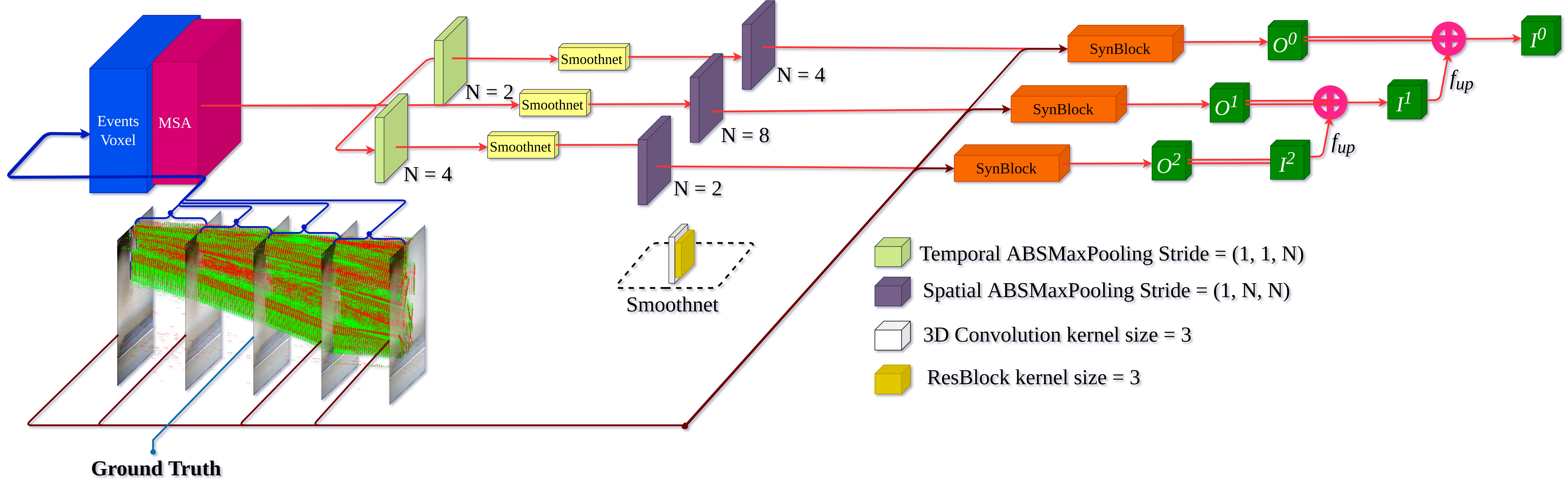}}
    \caption{Overview of the proposed method. Our method consists of three main parts. First events are represented with voxel grids. Then the voxel grids go through the SmoothNet and spatio-temporal absPooling layers. Finally, the constructed feature maps are fused with RGB frames by deformable convolutions that are embedded in parallel SynBlocks.} 
    \label{archFigure}
\end{figure*}

Our aim is insertion of a non-existent image sample into a video sequence that is composed of four images and four event intervals. Let the interpolated image be denoted by \(I_0\). Assume the input images are denoted as \(I_{-2}\), \(I_{-1}\), \(I_{+1}\), and \(I_{+2}\), which represent the successive original keyframes. The input event intervals are indicated by \(E_{-2 \to -1}\), \(E_{-1 \to 0}\), \(E_{0 \to 1}\), and \(E_{1 \to 2}\) which are described in terms of voxels whose dimensions are $4 \times N_{TB} \times H \times W$ ($N_{TB}$: number of time bins, $H$: height, $W$: width).

The block diagram of the proposed system is presented in Figure \ref*{archFigure}. The proposed method is composed of two stages. The events are converted into voxel representation in the first stage and passed to the multi-head self-attention (MSA) block. The resultant vector is passed through a series of blocks in which pooling operations are performed. As a slight difference from the literature, before applying maximum pooling, the absolute values of the vector elements are evaluated. However, after the pooling operation, the polarities of the vector elements are preserved; this stage is denoted as \textit{absPooling}. 

As a result of the first stage, three feature vectors are obtained, each of which encodes information coming from different resolutions. By the help of such pooling operations, the objects that are moving fast are predicted better, since using images at different resolutions makes it easier to associate fast-moving objects between consecutive images. In order to make this association even better, both the temporal and spatial domain absPooling operations are used. 

In the second stage, the feature vectors due to events resulting from the first step are provided as input to three SynBlocks (Synthesis Blocks \cite{Shi_2022_CVPR}) to fuse with RGB images. Every SynBlock takes four RGB frames and one event feature vector. Each SynBlock outputs an intermediate frame downscaled with \(l \in {0,1,2}\). The outputs of these three SynBlocks are combined to create the final output of our algorithm, which is shown in Figure \ref*{archFigure}, where \(f_{up}\) stands for bilinear upscaling, and \(O^l\) is an output of a SynBlock. Such a downscaling/upscaling structure helps the network to model trajectories of fast-moving objects better.

\textbf{Event-Volume Representation and Input Stage.} 
Event data coming from an event-based camera is asynchronous and sparse. The data is usually in $\mu s$ precision in time and streamed as a series of events (packages of $[x,y, timestamp, polarity]$). In a voxel representation, timestamps in the voxel grid are accumulated according to the number of time bins. The voxels for our case have dimensions of $4 \times N_{TB} \times H \times W$ and a similar approach as in \cite{Tulyakov_2021_CVPR} is followed for the generation of the voxels. The first event interval \(E_{-2 \to -1}\) and the second event interval \(E_{-1 \to 0}\) are used directly, but the third \(E_{0 \to 1}\) and the fourth \(E_{1 \to 2}\) event intervals are utilized as time-reversed. The first stage consists of two aforementioned absPooling layers. The first pooling layer calculates absPooling operation in the temporal dimension, whereas the second pooling layer calculates the same operation on its spatial counterpart. The multi-head self-attention mechanism (MSA), whose number of heads is 16, calculates attention values over the temporal dimension. These results are fed into SmoothNet blocks where these blocks simply include convolutional and ResNet layers, as in \cite{Shi_2022_CVPR}. This step is performed for the spatial smoothness of event feature vectors before the second (spatial) pooling operation. Therefore, the resultant three feature vectors, which encode different event-depth information, are processed in the first stage of the proposed method.

\begin{figure*}[htbp]
    \centering{\includegraphics[width=\linewidth]{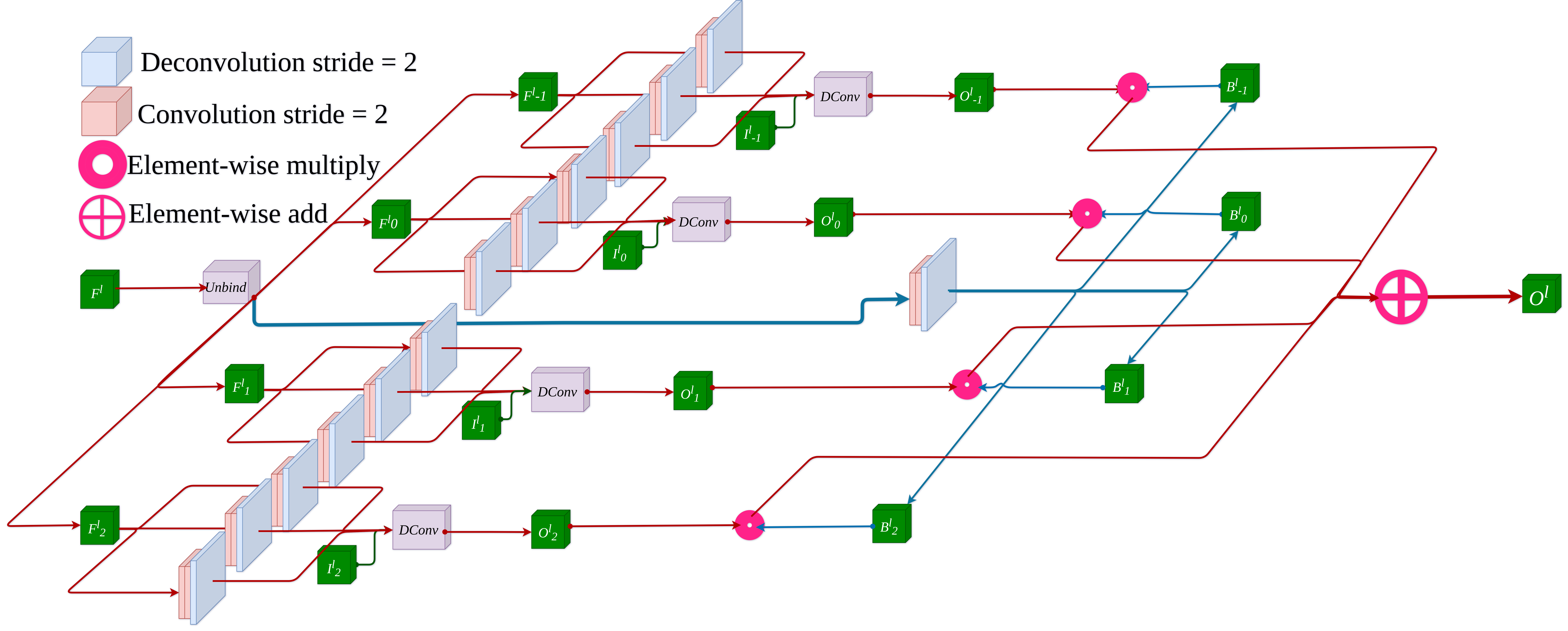}}
    \caption{The general structure of SynBlocks. A SynBlock takes four RGB frames and one event feature vector. Using these inputs, the SynBlocks fuse the event-based information and RGB images using kernels by utilizing deformable convolutions. }
    \label{figSynBlock}
\end{figure*}

\textbf{Fusion with Frames.}
SynBlocks are adapted from \cite{Shi_2022_CVPR} and presented in Figure \ref*{figSynBlock}. SynBlocks are the main synthesis blocks that can be considered as a multi-frame generalization of \cite{Lee2020AdaCoFAC}. The input part of a SynBlock starts from unbinding operation that the feature vectors \(F^l\) are divided into four in the temporal dimension. Each of these four temporal parts passes three parallel convolutional blocks generating four sisters of convolutional kernels. Each convolutional block constructs these convolutional kernels, and these kernels are created by 2D convolutions operating over event feature maps that produce deformable kernels for images. In other words, these three kernels are weights \(W_t^l \in  R^{K \times H \times W} \), horizontal offsets, \(\alpha_t^l \in  R^{K \times H \times W} \) and vertical offsets \(\beta_t^l \in  R^{K \times H \times W} \), where \(K\) stands for the number of sampling locations of every kernel. Using these predicted kernels, the result of the deformable convolution is reached, as in \cite{Lee2020AdaCoFAC}:
\[
O_t^l(x,y) =  \sum_{n = 1}^{K} W_t^l (n,x,y) I_t^l (x+\alpha_t^l(n, x, y),y+\beta_t^l(n, x, y))      
\]

Therefore, RGB frames, \(I^l_t\), are fused by the help of feature vectors, $W_t^l$, $\alpha_t^l$ and $\beta_t^l$  that are obtained from convolutional blocks to reach deformable convolution block operation \cite{Lee2020AdaCoFAC}. In the last step of a SynBlock,  \(O^l_t\) is blended with learned masks. The masks are acquired by using standard convolutional neural networks on the feature map that is the concatenation of the SynBlock input. The result of a SynBlock is obtained by elementwise multiplication,
\[
O^l = \sum_{t} B_t^l \cdot   O_t^l   
\]

Finally, \(O^l_t\) vectors are elementwise multiplied with the vector output of standalone convolutional block \(B^l_t\).

As each SynBlock generates the output at scale \(l \in {0,1,2}\) the intermediate image is determined as,
\[
\hat{I}^l_0 = f_{up}(\hat{I}_0) + O^l     
\] where \(f_{up}\) stands for bilinear upscaling, and \(O^l\) is an output of a SynBlock. This operation concludes the fusion of frames with event feature maps and constructs intermediate frames. 

The proposed method is a relatively simple yet effective approach for the utilization of additional temporal information supplied by the events. Differing from \cite{Shi_2022_CVPR}, the initial multi-head self-attention block in the input layer is small-sized and significantly reduces the model size, while still modeling trajectories by the help of precise event information. On the other hand, it should be pointed out that the voxels are supplied into the network in a dense manner, which means the sparsity of the events is not fully exploited in our work. 

\section{Experiments and Results}
\subsection{Implementation Details}

All of our work on deep learning is performed by using PyTorch \cite{PyTorch} framework. The training has performed for 36 epochs, where the learning rate, starting from $0.0008$, is halved every eight epochs. We have used AdaMax optimizer \cite{AdaMax} with $\beta_1=0.9, \beta_2=0.999$ and the training batch size is set as $6$. The training has been executed on a workstation with two 2080TI GPUs. Due to the memory requirements, the training has been performed with images and associated events whose resolution is $256 \times 256$. We have tested and compared the proposed method with both full-scaled and downscaled frames with $256 \times 256$ resolution, since the obtained quantitative results can differ significantly with different sized images.

\subsection{Quantitative Results} All the algorithms are tested on BS-ERGB \cite{Tulyakov_2022_CVPR} dataset.  It should be noted that the test results of \cite{Tulyakov_2021_CVPR} are replicated from \cite{Tulyakov_2022_CVPR} directly. For the algorithms in \cite{Shi_2022_CVPR} and \cite{Kalluri2020FLAVRFV}, we have conducted experiments by using the codes provided by the authors and the listed the best results. 

\begin{figure}[htbp]
    \centerline{\includegraphics[width=\linewidth]{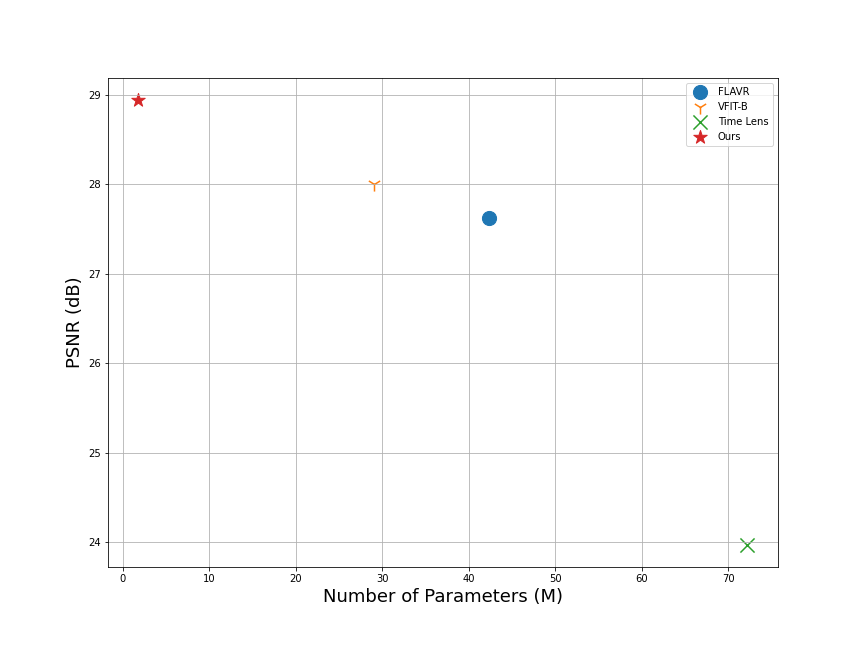}}
    \caption{Comparison of the performance of different algorithms (FLAVR \cite{Kalluri2020FLAVRFV}, VFIT-B \cite{Shi_2022_CVPR}, Time Lens \cite{Tulyakov_2021_CVPR}, Time Lens++ \cite{Tulyakov_2022_CVPR}, and our proposed method):  number of parameters vs. PSNR values for BS-ERGB [Full Scale] \cite{Tulyakov_2022_CVPR} dataset.}
    \label{numParam2}
\end{figure}

\begin{figure}[htbp]
    \centerline{\includegraphics[width=\linewidth]{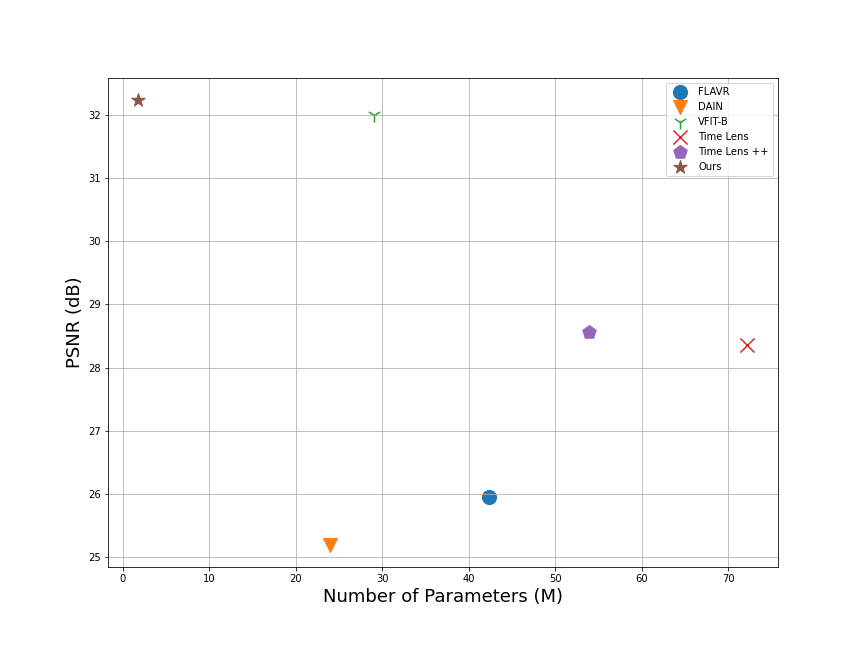}}
    \caption{Comparison of the performance of different algorithms (FLAVR \cite{Kalluri2020FLAVRFV}, VFIT-B \cite{Shi_2022_CVPR}, Time Lens \cite{Tulyakov_2021_CVPR}, Time Lens++ \cite{Tulyakov_2022_CVPR}, and our proposed method):  number of parameters vs. PSNR values for BS-ERGB \cite{Tulyakov_2022_CVPR} dataset.}
    \label{numParam}
\end{figure}

\begin{table}[htbp]
    \caption{Comparison of our proposed method in BS-ERGB dataset in low resolution}
    \begin{center}
        \begin{tabular}{|c|c|c|c|c|}
        \hline
        \textbf{Method} & \textbf{Input} &\textbf{\shortstack{ \#Parameters\\(M)}}& \textbf{PSNR (dB)}& \textbf{SSIM} \\
        \hline
        FLAVRFV \cite{Kalluri2020FLAVRFV}&Frames& 42.4 &31.72 & 0.9469 \\
        \hline
        VFIT \cite{Shi_2022_CVPR} &Frames& 29.0  & 32.08&  0.9449\\
        \hline
        Timelens \cite{Tulyakov_2021_CVPR} &\shortstack{Frames\\Events}& 72.2 & 28.36 & 0.9320 \\
        \hline
        Timelens++ \cite{Tulyakov_2022_CVPR} &\shortstack{Frames\\Events}& 53.9 & 28.56 &  - \\
        \hline
        \textbf{Ours} &  \textbf{\shortstack{Frames\\Events}} & \textbf{2.07} & \textbf{32.23} & \textbf{0.9581} \\
        \hline
        \end{tabular}
    
    \label{tabComp256}
\end{center}
\end{table}

\begin{table}[htbp]
    \caption{Comparison of our proposed method in BS-ERGB dataset in full scale}
    \begin{center}
        \begin{tabular}{|c|c|c|c|c|}
        \hline
        \textbf{Method} & \textbf{Input} &\textbf{\shortstack{ \#Parameters\\(M)}}& \textbf{PSNR (dB)}& \textbf{SSIM} \\
        \hline
        FLAVRFV \cite{Kalluri2020FLAVRFV}&Frames& 42.4 &27.642 & 0.8729 \\
        \hline
        VFIT \cite{Shi_2022_CVPR} &Frames& 29.0  & 28.00 &  0.8767\\
        \hline
        Timelens \cite{Tulyakov_2021_CVPR} &\shortstack{Frames\\Events}& 72.2 & 23.97 & 0.7838 \\
        \hline
        Timelens++ \cite{Tulyakov_2022_CVPR} &\shortstack{Frames\\Events}& 53.9 & - &  - \\
        \hline
        \textbf{Ours} & \textbf{\shortstack{Frames\\Events}} & \textbf{2.07} & \textbf{29.04} & \textbf{0.8771} \\
        \hline
        \end{tabular}
    \label{tabCompFull}
\end{center}
\end{table}

Tables \ref*{tabComp256} and \ref*{tabCompFull} present the quantitative results for BS-ERGB dataset in full-scale and down-scaled to $256 \times 256$, where peak signal-to-noise ratio (PSNR) and structural similarity index measure (SSIM) values are used for comparison. Due to the rich temporal information stored in event voxels and the utilization of deformable convolutions, the proposed method clearly outperforms the current state-of-the-art event-based method by a significant margin of $5$ dB PSNR on BS-ERGB dataset \cite{Tulyakov_2022_CVPR}. Moreover, the proposed method surpasses the state-of-art frame-only method by $1$ dB, while utilizing only $6.9 \%$ of its original model size. As plotted in \ref{numParam2} ,and \ref{numParam} the proposed method significantly reduces the number of parameters  compared to both event-based and frame-only methods. That is, the proposed method promises faster and more feasible local computation option.

\subsection{Qualitative Results}
\begin{figure*}[!t]
	\begin{center}
	\begin{minipage}[h]{0.16\linewidth}
		\centering
		\includegraphics[width=\linewidth]{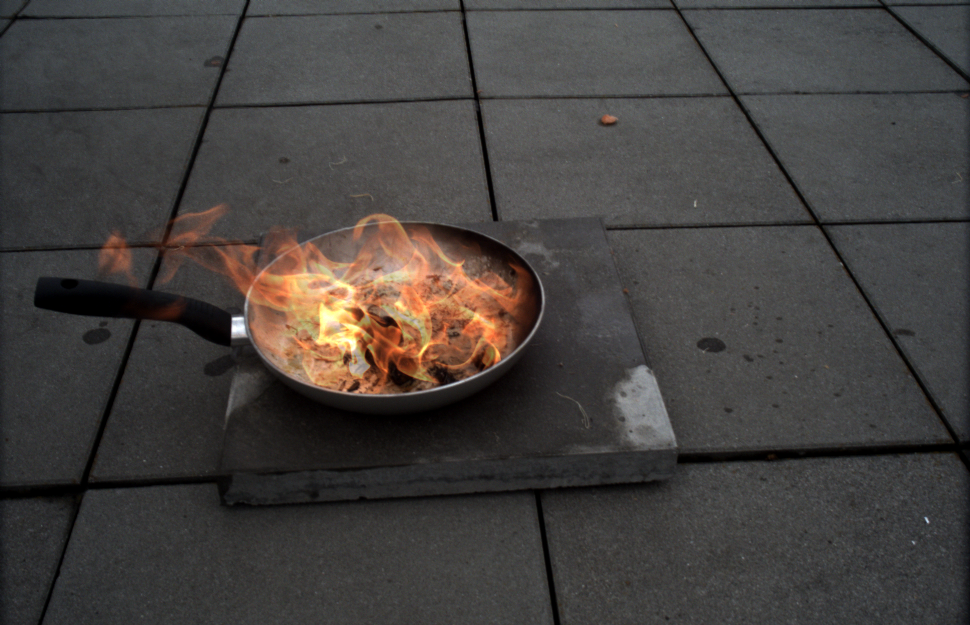}
	\end{minipage}
	\begin{minipage}[h]{0.16\linewidth}
		\centering
		\includegraphics[width=\linewidth]{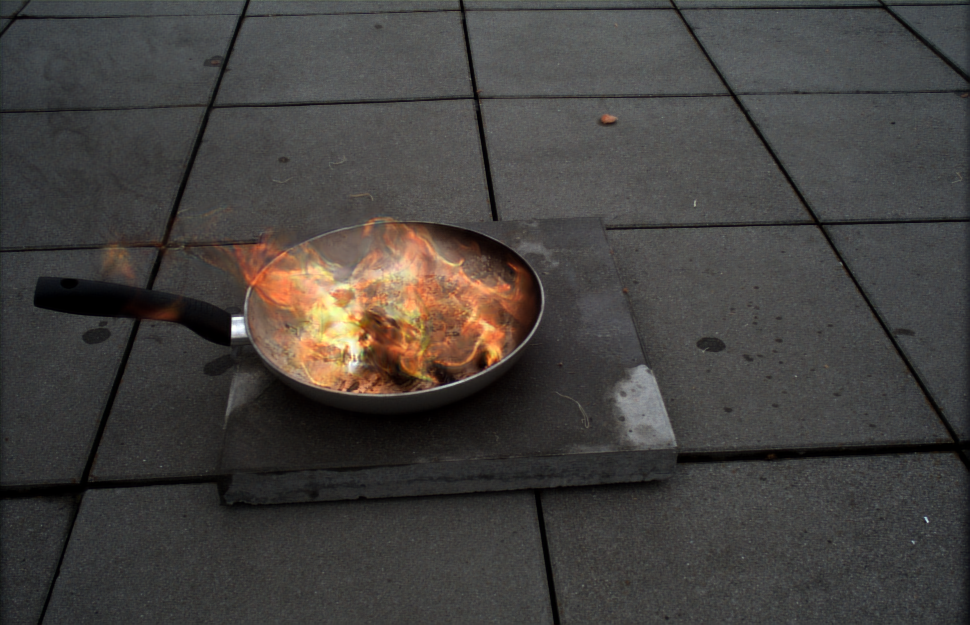}
	\end{minipage}
	\begin{minipage}[h]{0.16\linewidth}
		\centering
		\includegraphics[width=\linewidth]{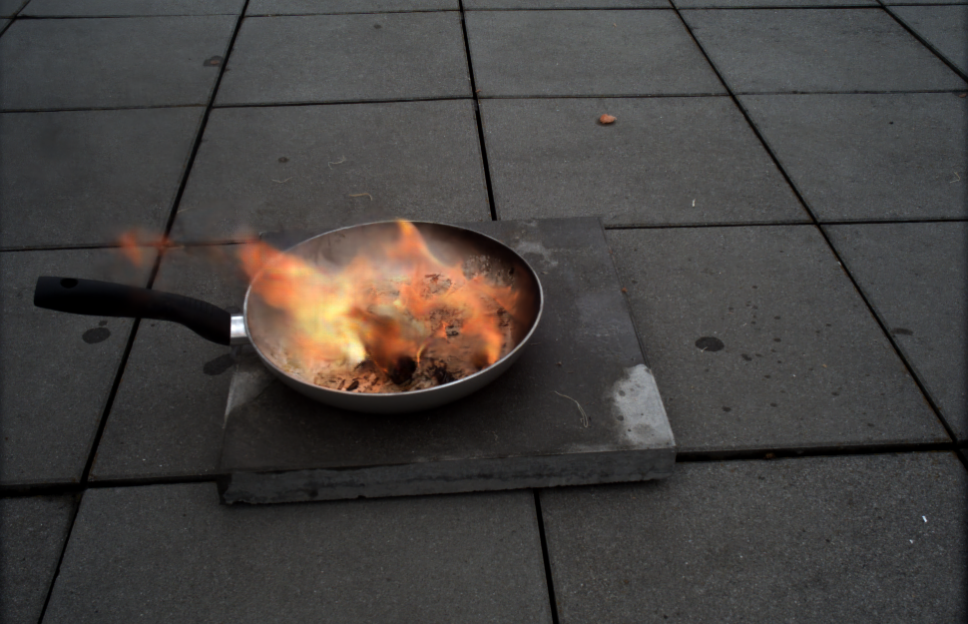}
	\end{minipage}
	\begin{minipage}[h]{0.16\linewidth}
		\centering
		\includegraphics[width=\linewidth]{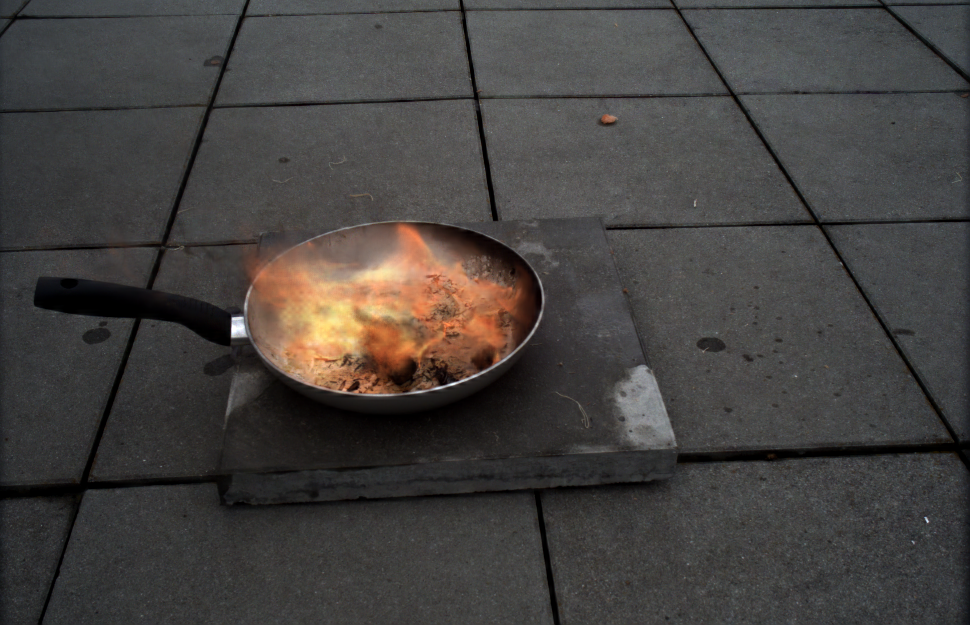}
	\end{minipage}
	\begin{minipage}[h]{0.16\linewidth}
		\centering
		\includegraphics[width=\linewidth]{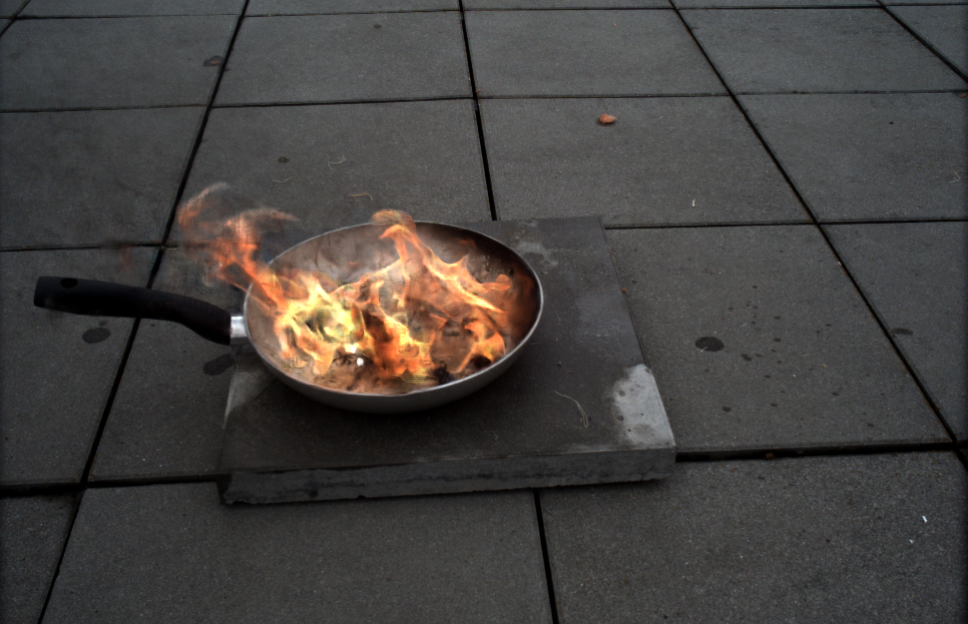}
	\end{minipage}
	\begin{minipage}[h]{0.16\linewidth}
		\centering
		\includegraphics[width=\linewidth]{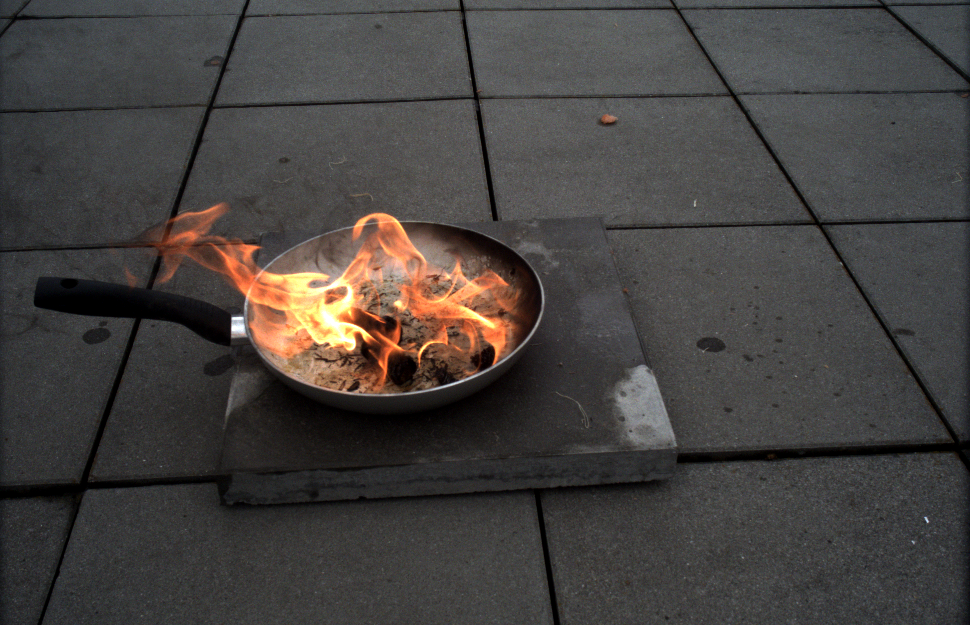}
	\end{minipage}
    \end{center}

    \begin{center}
	\begin{minipage}[h]{0.16\linewidth}
		\centering
		\includegraphics[width=\linewidth]{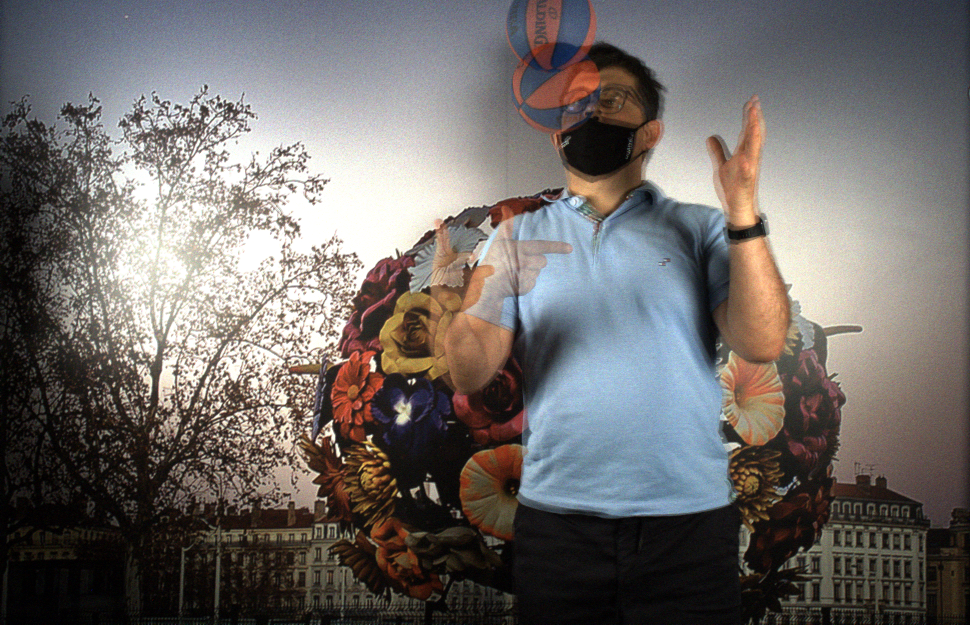}
		\scriptsize{Overlayed}
	\end{minipage}
	\begin{minipage}[h]{0.16\linewidth}
		\centering
		\includegraphics[width=\linewidth]{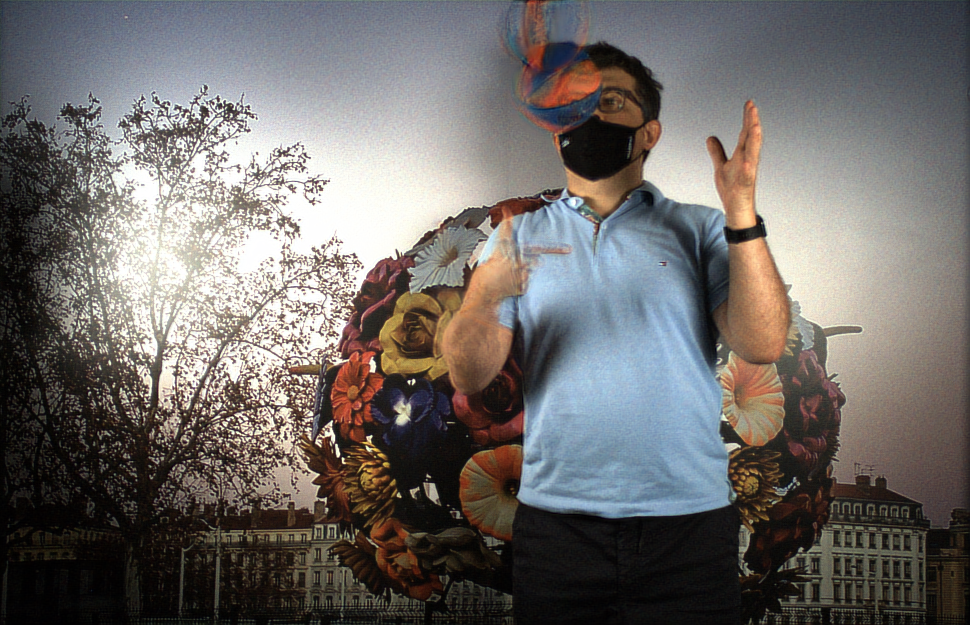}
		\scriptsize{Time Lens}
	\end{minipage}
	\begin{minipage}[h]{0.16\linewidth}
		\centering
		\includegraphics[width=\linewidth]{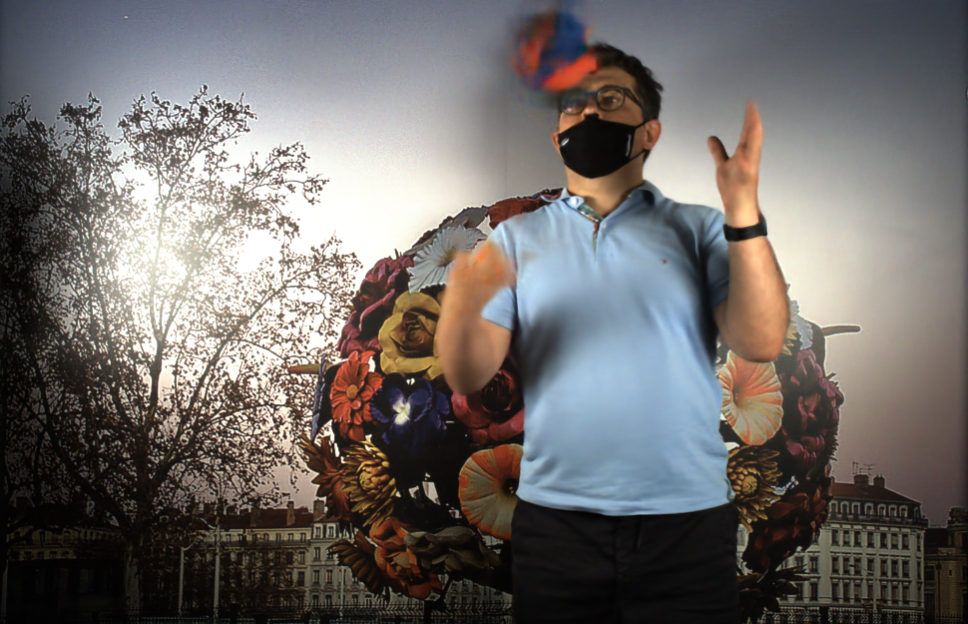}
		\scriptsize{FLAVR}
	\end{minipage}
	\begin{minipage}[h]{0.16\linewidth}
		\centering
		\includegraphics[width=\linewidth]{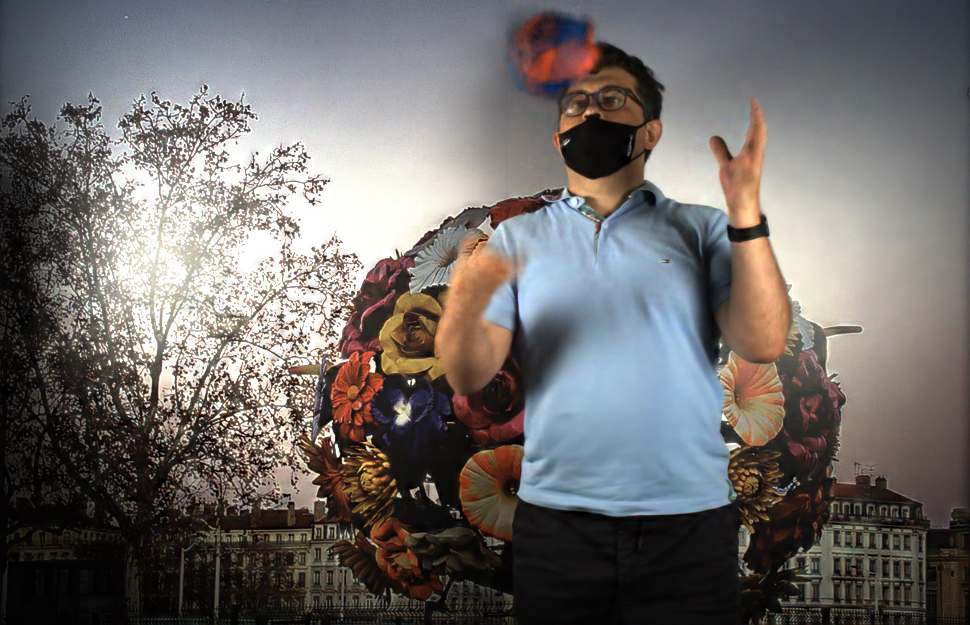}
		\scriptsize{VFIT}
	\end{minipage}
	\begin{minipage}[h]{0.16\linewidth}
		\centering
		\includegraphics[width=\linewidth]{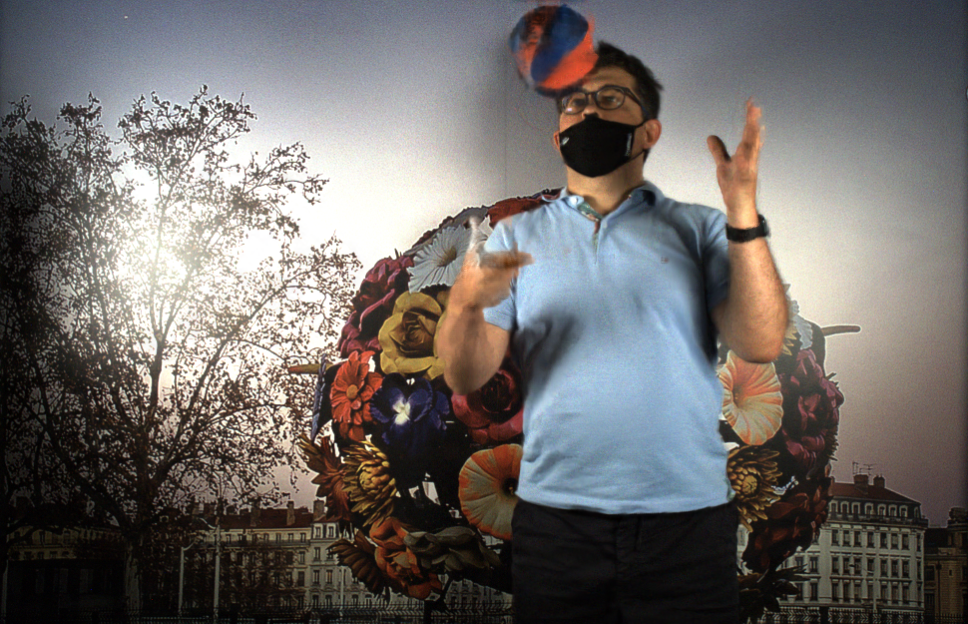}
		\scriptsize{Ours}
	\end{minipage}
	\begin{minipage}[h]{0.16\linewidth}
		\centering
		\includegraphics[width=\linewidth]{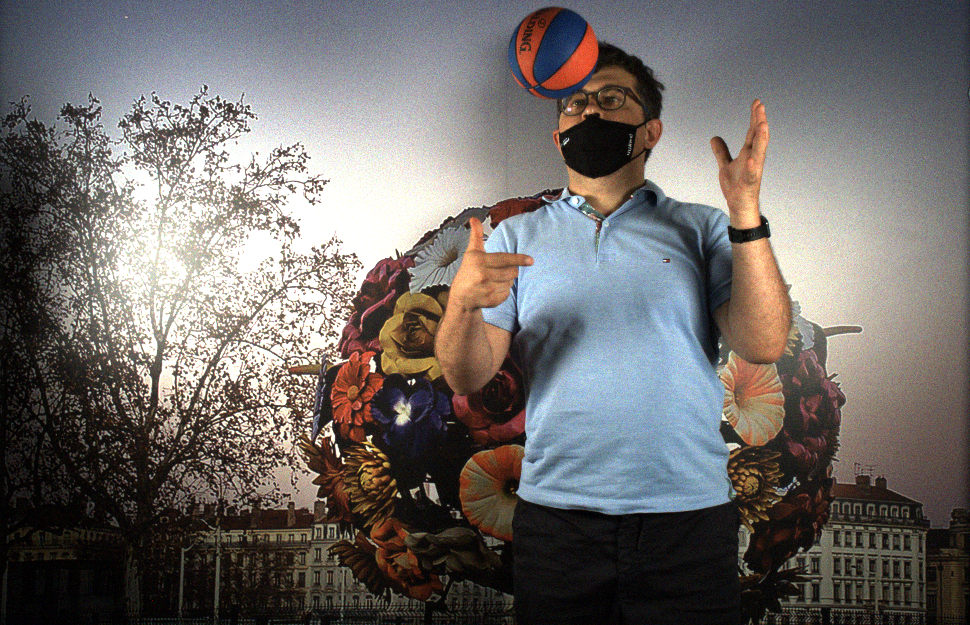}
		\scriptsize{GT}
	\end{minipage}
    \end{center}
\vspace{-1mm}
	\caption{{Qualitative comparisons} against the state-of-the-art video interpolation algorithm. Our method is less prone to blur and ghosting effects. Thus, it provides more realistic increments on frame-rates of real videos. 
    }
	\vspace{-2mm}
	\label{fig:qualitative_comparisons}
\end{figure*}

\begin{figure*}[!t]
	\begin{center}
	\begin{minipage}[h]{0.16\linewidth}
		\centering
		\includegraphics[width=\linewidth]{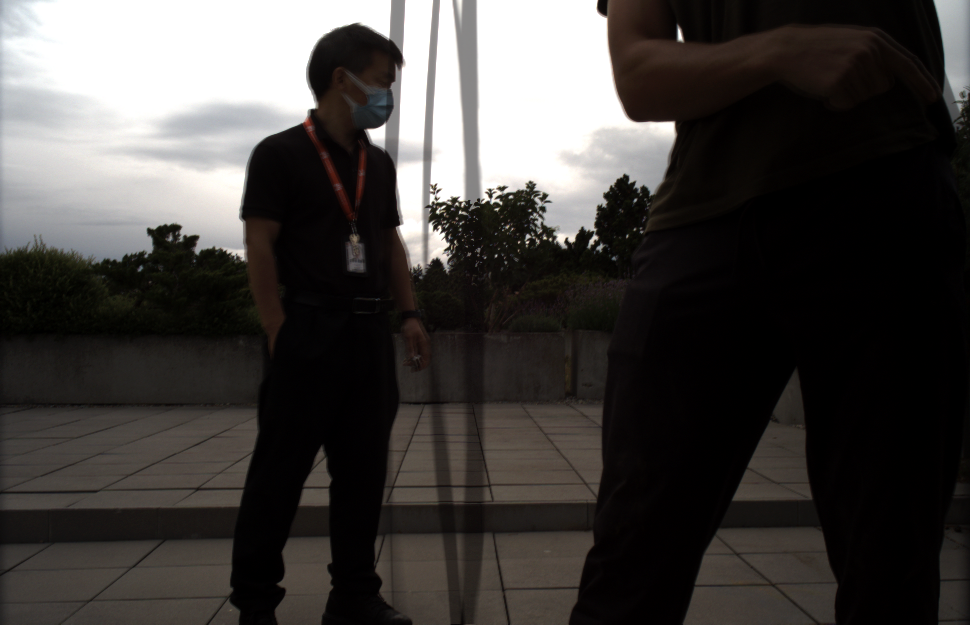}
	\end{minipage}
	\begin{minipage}[h]{0.16\linewidth}
		\centering
		\includegraphics[width=\linewidth]{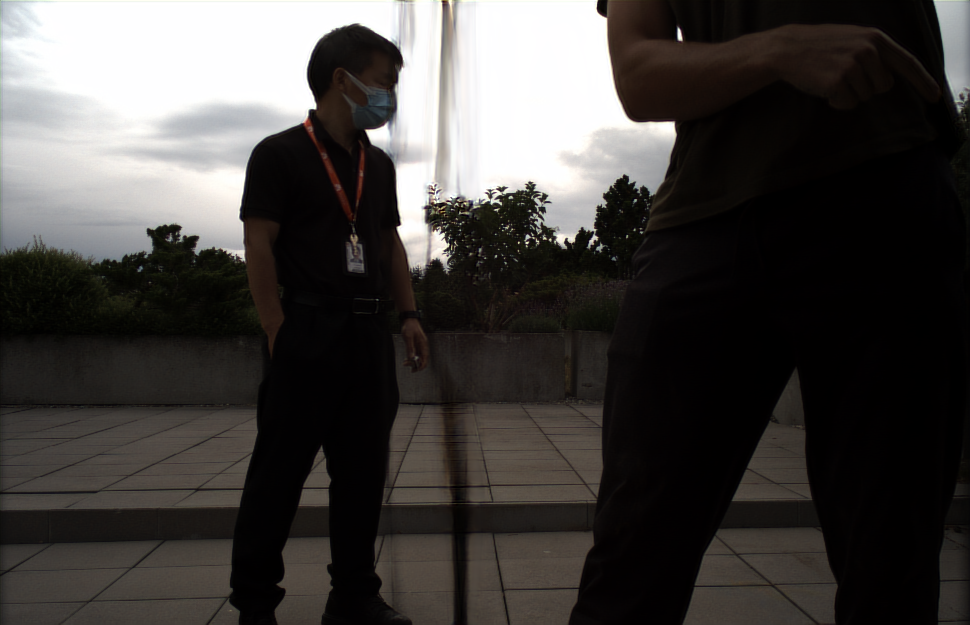}
	\end{minipage}
	\begin{minipage}[h]{0.16\linewidth}
		\centering
		\includegraphics[width=\linewidth]{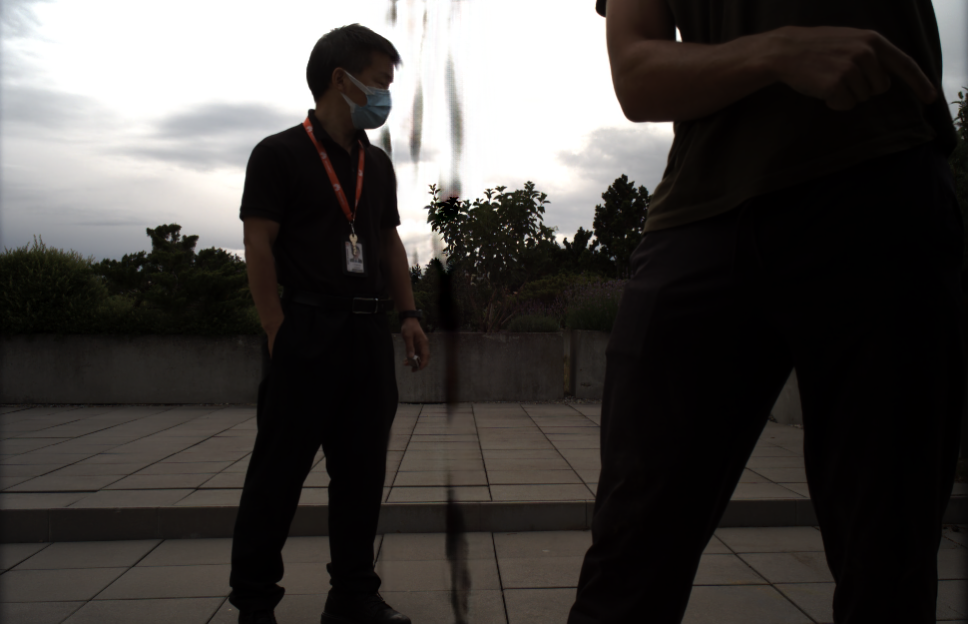}
	\end{minipage}
	\begin{minipage}[h]{0.16\linewidth}
		\centering
		\includegraphics[width=\linewidth]{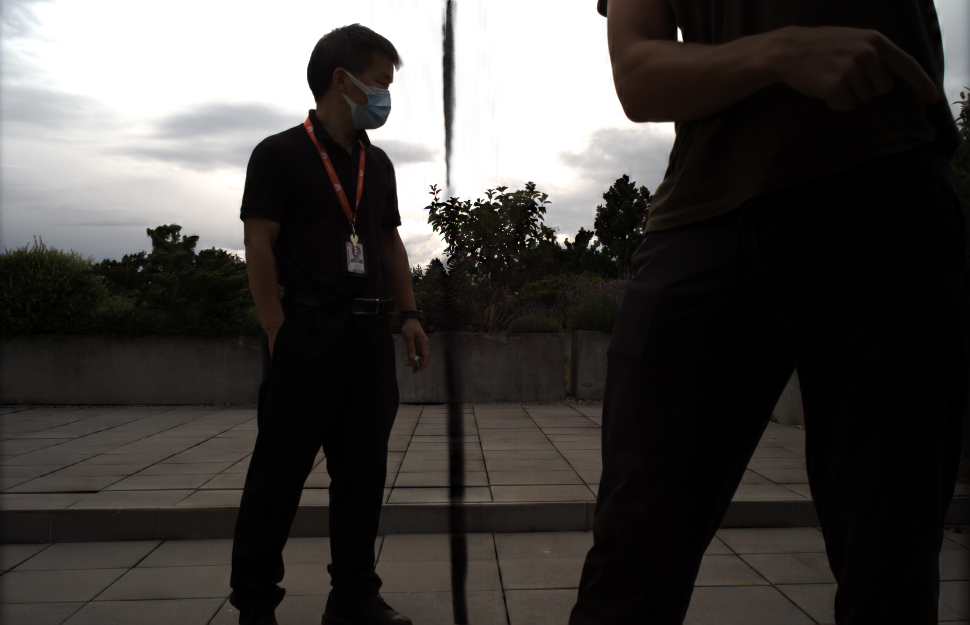}
	\end{minipage}
	\begin{minipage}[h]{0.16\linewidth}
		\centering
		\includegraphics[width=\linewidth]{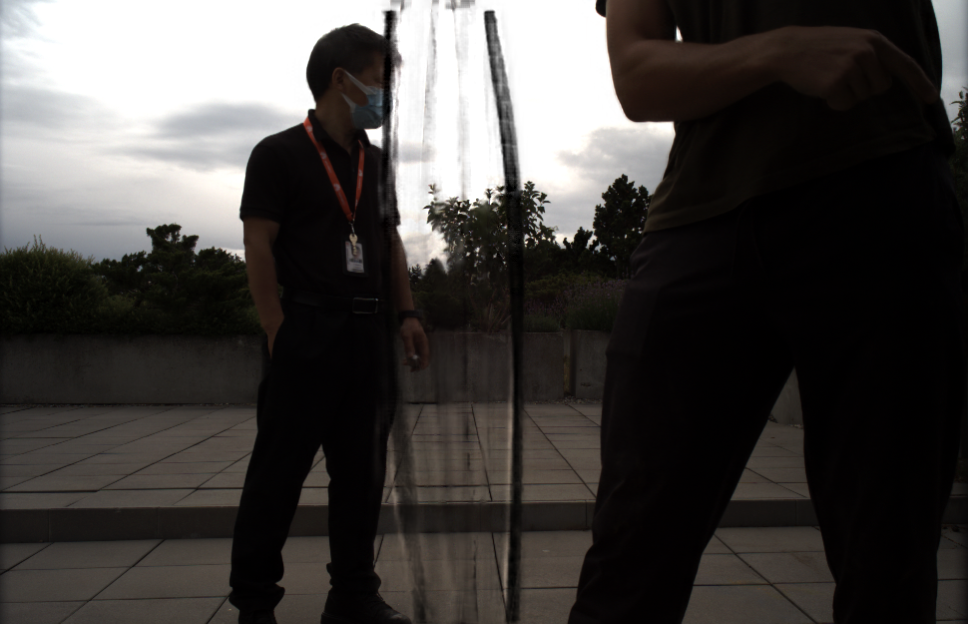}
	\end{minipage}
	\begin{minipage}[h]{0.16\linewidth}
		\centering
		\includegraphics[width=\linewidth]{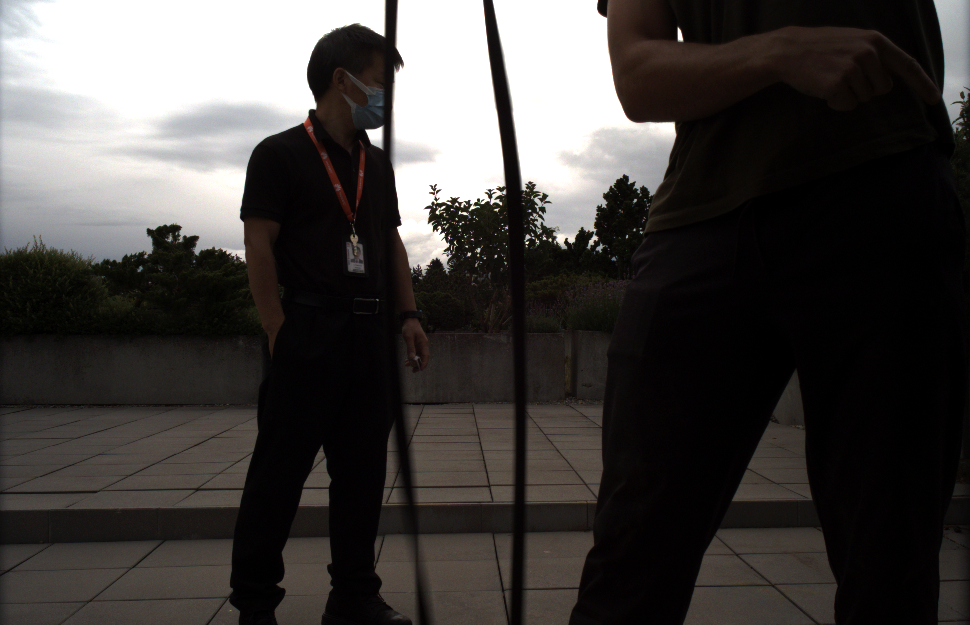}
	\end{minipage}
    \end{center}

    \begin{center}

	\begin{minipage}[h]{0.16\linewidth}
		\centering
		\includegraphics[width=\linewidth]{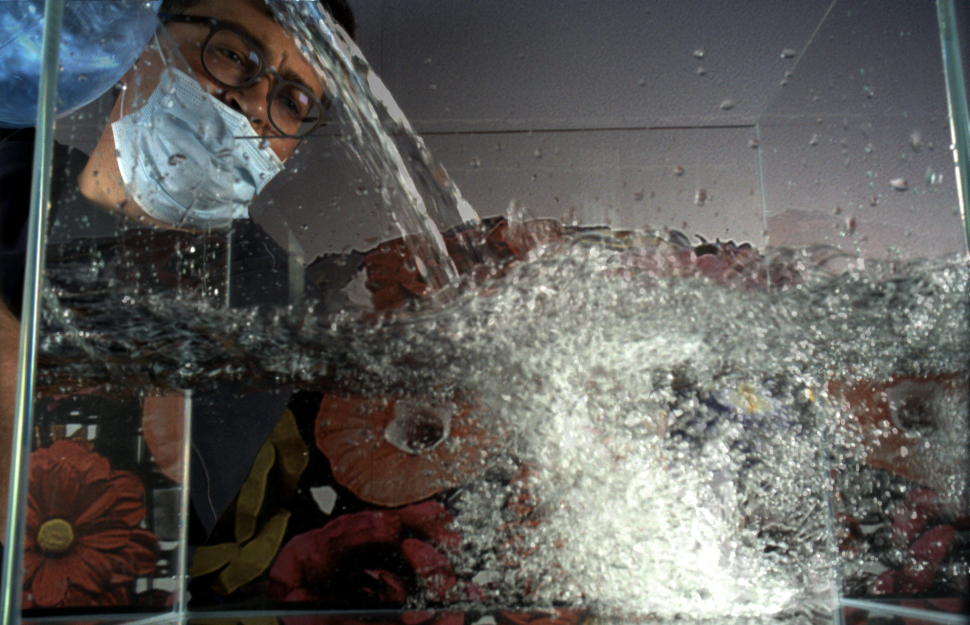}
		\scriptsize{Overlayed}
	\end{minipage}
	\begin{minipage}[h]{0.16\linewidth}
		\centering
		\includegraphics[width=\linewidth]{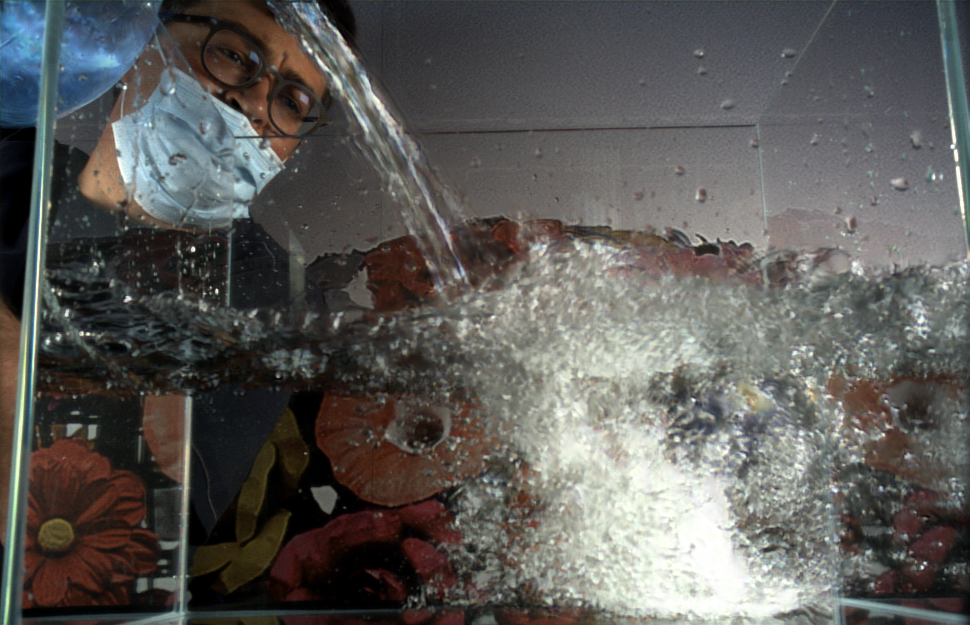}
		\scriptsize{Time Lens}
	\end{minipage}
	\begin{minipage}[h]{0.16\linewidth}
		\centering
		\includegraphics[width=\linewidth]{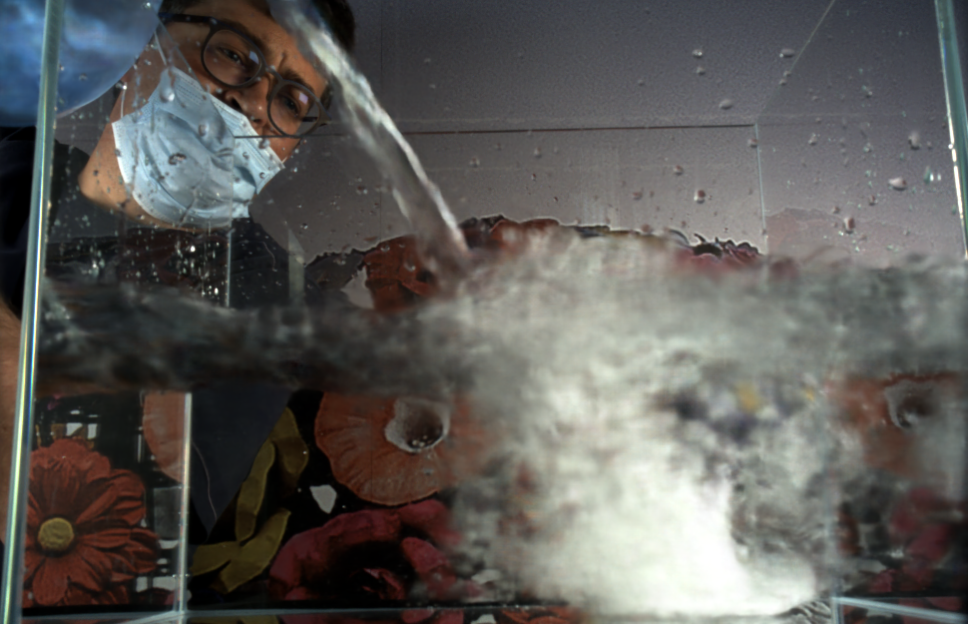}
		\scriptsize{FLAVR}
	\end{minipage}
	\begin{minipage}[h]{0.16\linewidth}
		\centering
		\includegraphics[width=\linewidth]{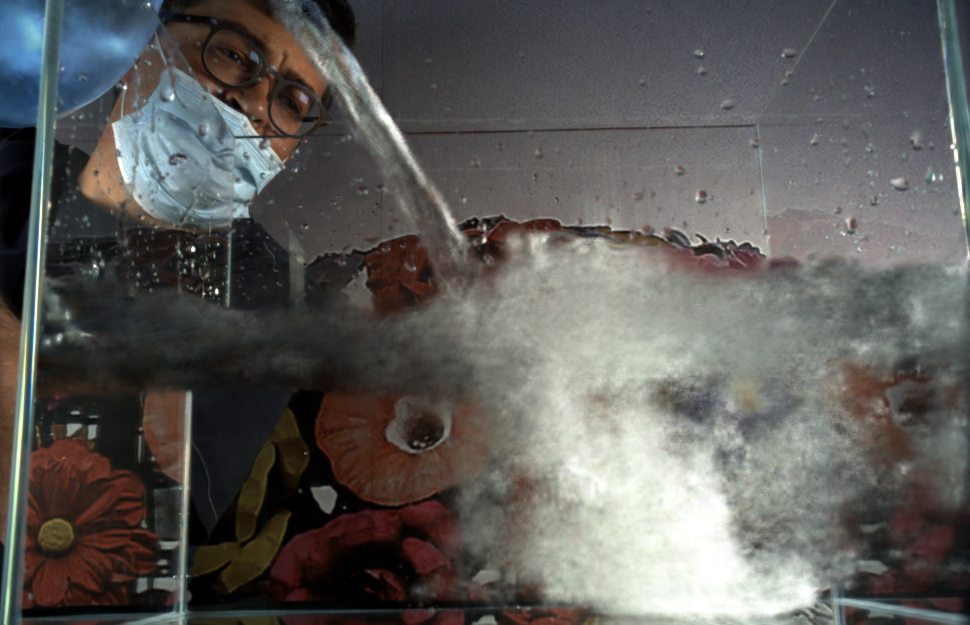}
		\scriptsize{VFIT}
	\end{minipage}
	\begin{minipage}[h]{0.16\linewidth}
		\centering
		\includegraphics[width=\linewidth]{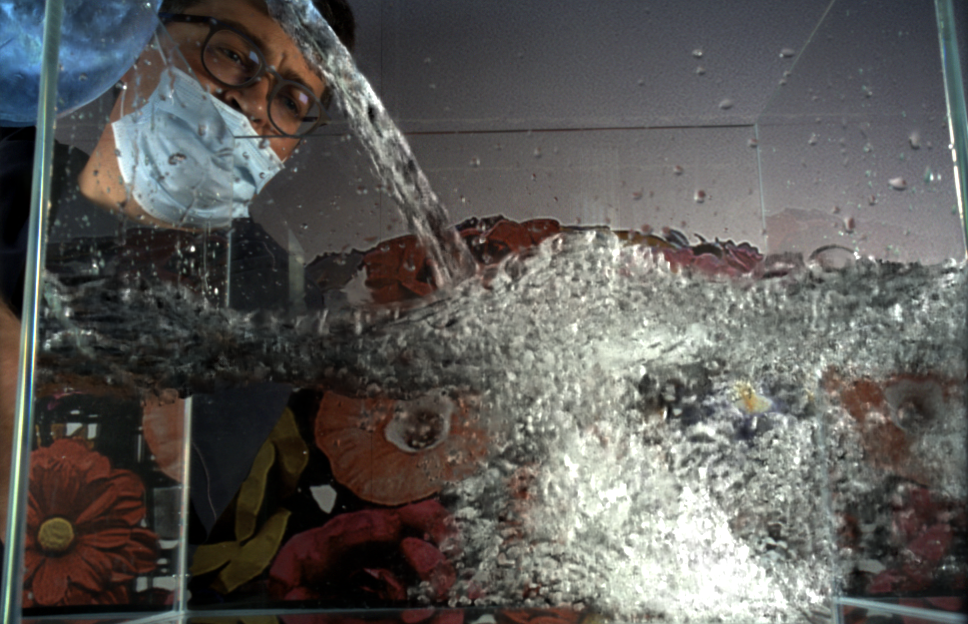}
		\scriptsize{Ours}
	\end{minipage}
	\begin{minipage}[h]{0.16\linewidth}
		\centering
		\includegraphics[width=\linewidth]{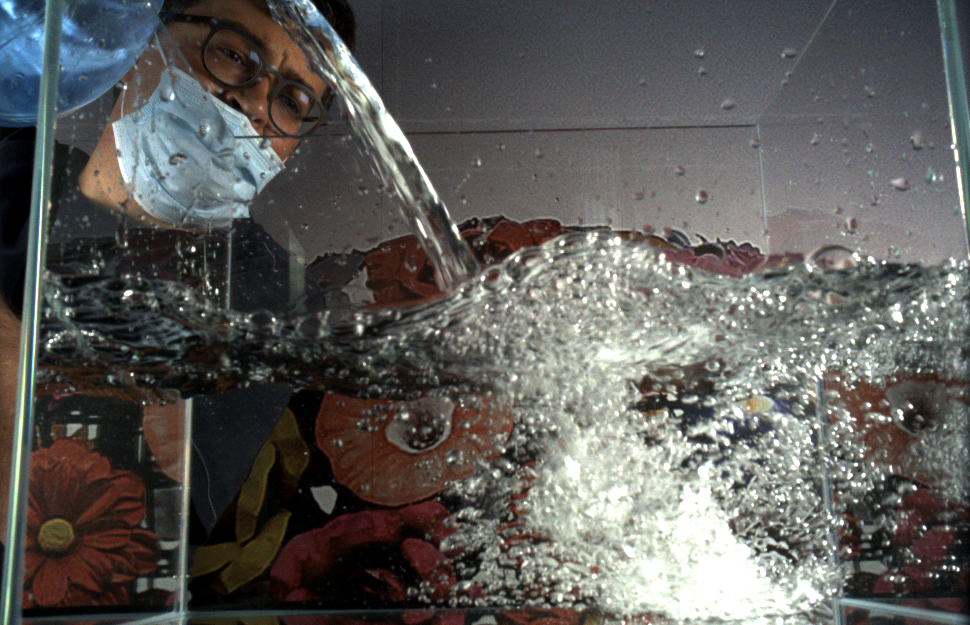}
		\scriptsize{GT}
	\end{minipage}
    \end{center}
\vspace{2mm}
	\caption{{Qualitative comparisons} against the state-of-the-art video interpolation algorithms.}
	\vspace{-2mm}
	\label{fig:qualitative_comparisons2}
\end{figure*}

\begin{figure*}

\begin{subfigure}{.24\textwidth}
      \centering
      \includegraphics[width=.8\linewidth]{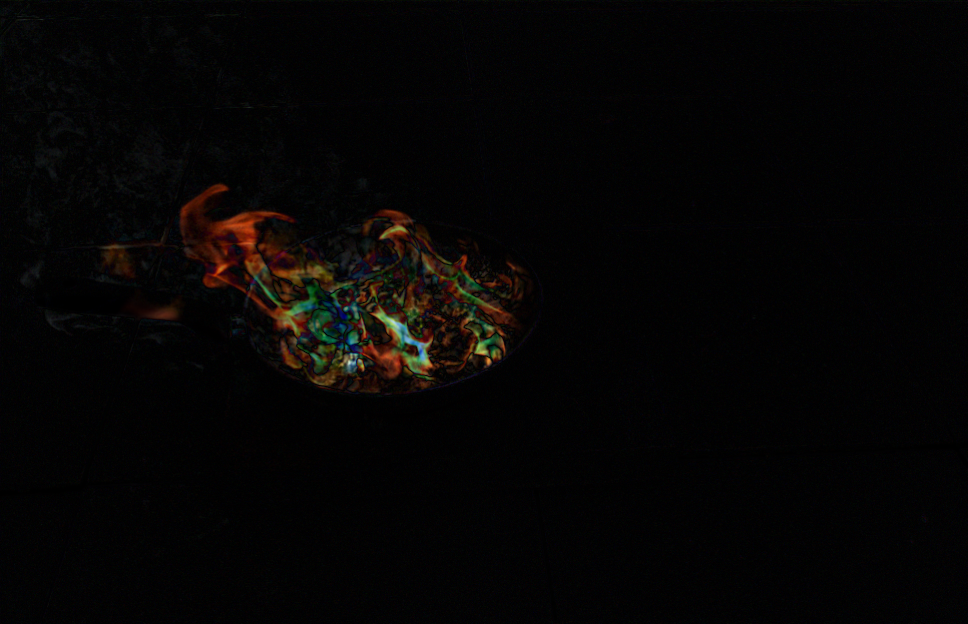}
      \caption{Time Lens}
      \label{fig:sfig2}
\end{subfigure}
\begin{subfigure}{.24\textwidth}
    \centering
    \includegraphics[width=.8\linewidth]{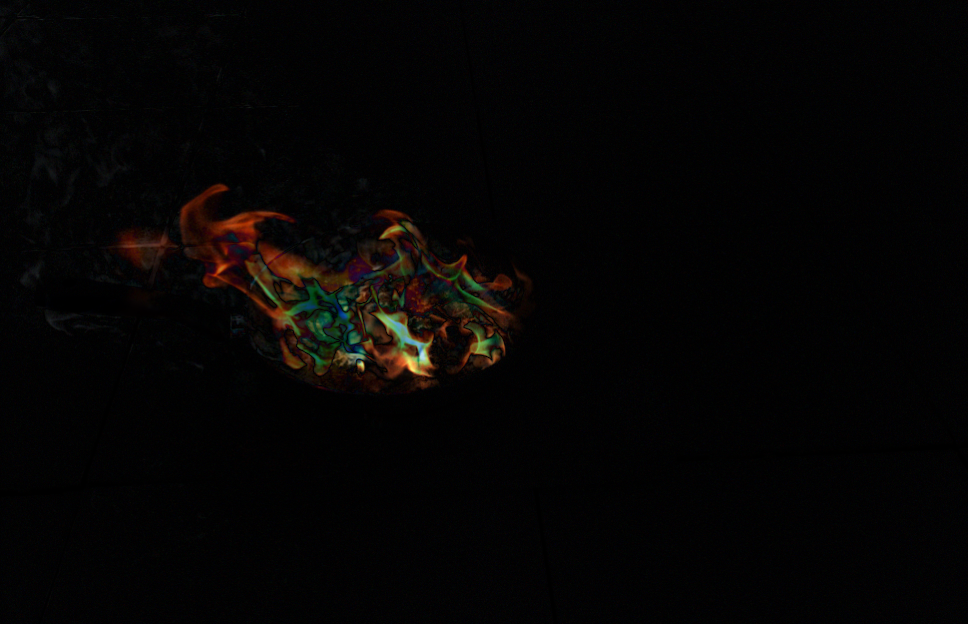}
    \caption{FLAVRFV}
    \label{fig:sfig4}
\end{subfigure}
\begin{subfigure}{.24\textwidth}
    \centering
    \includegraphics[width=.8\linewidth]{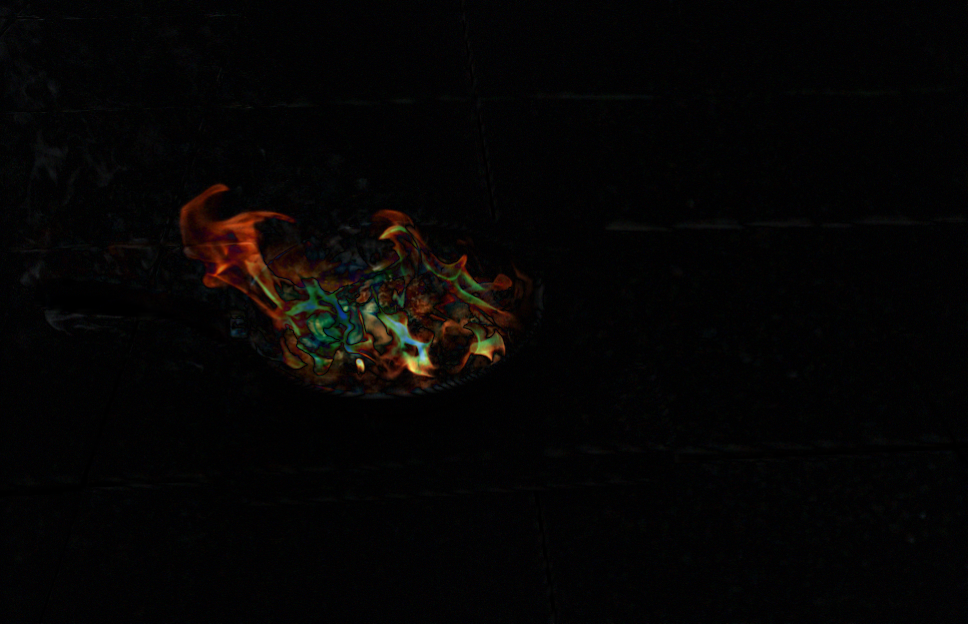}
    \caption{VFIT}
    \label{fig:sfig5}
\end{subfigure}
\begin{subfigure}{.24\textwidth}
    \centering
    \includegraphics[width=.8\linewidth]{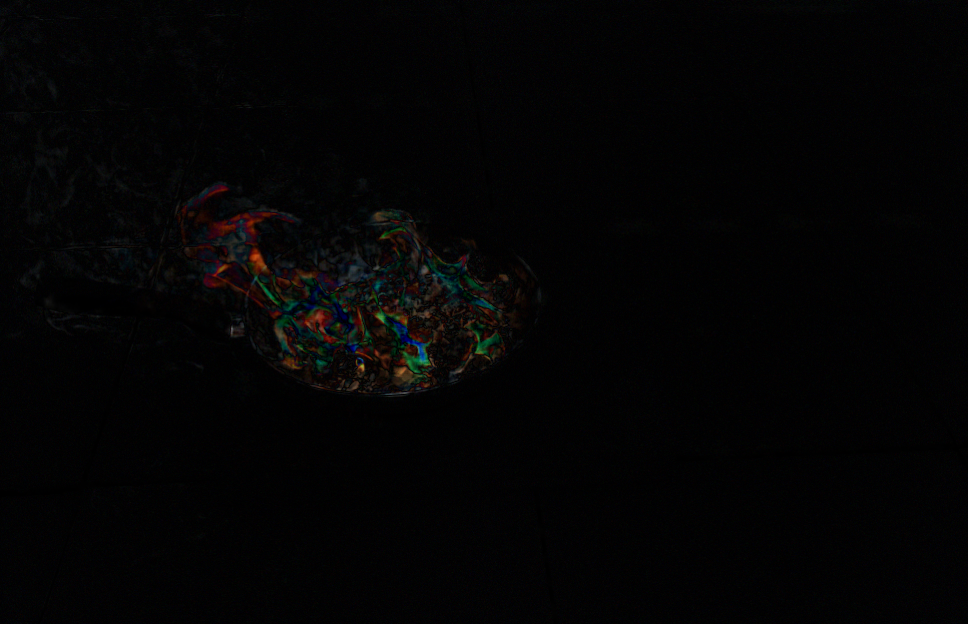}
    \caption{Ours}
    \label{fig:sfig6}
\end{subfigure}

\caption{Pixel-wise differences of the fire sample.}
\label{fig:fig}
\end{figure*}

We have also compared the proposed method qualitatively by other frame-only and event-based methods with the full-scale BS-ERGB \cite{Tulyakov_2022_CVPR} dataset. Typical examples of the outcomes are illustrated in Figures \ref*{fig:qualitative_comparisons} and \ref*{fig:qualitative_comparisons2}. As it can be observed, our method provides crispier interpolated frames. The \ref{fig:fig} shows that the proposed method achieves more accurate results in all three channels as the other methods have more heterogeneous pixel-wise difference maps. Furthermore, the objects that have complex motion with non-linear trajectories are estimated more accurately compared to other techniques.  

\subsection{Ablation Studies}
We have conducted a series of ablation studies to better comprehend the proposed strategy. First, in order to validate the effectiveness of the multi-head self-attention mechanism, we have discarded this mechanism during simulations. Additionally, we have also analyzed the effect of Sep-STS layers proposed in \cite{Shi_2022_CVPR} on event voxels. 
\subsubsection{Results for Events without Attention Mechanism}
To investigate the significance of the attention mechanism, we have trained a network without the mechanism in the input stage.

The quantitative results in Table \ref*{tabCompWoMSA} indicate that the attention mechanism in the input stage of the network increases the performance by $0.59$dB. This result is important, since the temporal attention mechanism intensifies the crucial elements of the voxels.

\begin{table}[htbp]
    \caption{Comparison in BS-ERGB dataset}
    \begin{center}
        \begin{tabular}{|c|c|c|}
        \hline
        \textbf{Method} & \textbf{PSNR (dB)}& \textbf{SSIM} \\
        \hline
        w/o MSA &31.64 & 0.9496 \\
        \hline
        \textbf{Ours} & \textbf{32.23} & \textbf{0.9581} \\
        \hline
        \end{tabular}
    \label{tabCompWoMSA}
\end{center}
\end{table}

\subsubsection{Result for Events with Sep-STS Layers}
The authors of \cite{Shi_2022_CVPR} have introduced Sep-STS blocks for the encoder part of the network adapted from \cite{Lee2020AdaCoFAC}. We have compared the encoder of \cite{Shi_2022_CVPR} with the first part of the proposed algorithm to investigate the performance of Sep-STS blocks by using events with both approaches. 

The comparative results of these experiments are given in Table \ref*{tabCompSepSTS}. Although the number of parameters presented by the Sep-STS blocks is relatively large, utilization of a simpler input stage yields better results due to the valuable temporal information in events. One can argue that Sep-STS might require more training data to yield better inference.

\begin{table}[htbp]
    \caption{Comparison in BS-ERGB dataset}
    \begin{center}
        \begin{tabular}{|c|c|c|}
        \hline
        \textbf{Method} & \textbf{PSNR (dB)}& \textbf{SSIM} \\
        \hline
        Sep-STS Encoder & 31.19 & 0.9436 \\
        \hline
        \textbf{Ours} & \textbf{32.23} & \textbf{0.9581} \\
        \hline
        \end{tabular}
    \label{tabCompSepSTS}
\end{center}
\end{table}

\section{Conclusion}

In this paper, we propose E-VFIA, a lightweight and high quality video frame interpolation algorithm that uses frames obtained from RGB cameras, as well as event information obtained from event cameras. The proposed method encodes event information with a multi-head self-attention mechanism and fuses the encoded information with frames under deformable convolution-based synthesis blocks. Consequently, the E-VFIA achieves $3.87$ dB and $5.07$ dB PSNR improvement over the state-of-the-art event-based method for low and high-resolution images, respectively. The model achieves this result by only $3.8 \%$ of its model size. It should be emphasized that based on qualitative comparisons, our method is less prone to blur and ghosting artifacts, whereas in some cases, suffers from the color inconsistency of the fast-moving objects. Finally, it can be argued that the proposed deformable convolutions strategy is a promising approach to fuse event information and visual appearance in any vision or robotics application, specifically VFI.

\addtolength{\textheight}{-8cm}   






\bibliography{IEEEbib/IEEEfull}


\end{document}